\documentclass[sigconf]{acmart}

\usepackage{booktabs} 
\usepackage{array}
\usepackage{subfigure}

\begin{document}

\copyrightyear{2017} 
\acmYear{2017} 
\setcopyright{acmcopyright}
\acmConference{MM '17}{October 23--27, 2017}{Mountain View, CA, USA}\acmPrice{15.00}\acmDOI{10.1145/3123266.3123328}
\acmISBN{978-1-4503-4906-2/17/10}

\fancyhead{}
\settopmatter{printacmref=false, printfolios=false}

\title{Hierarchical Recurrent Neural Network for Video Summarization}	

\author{Bin Zhao}
\affiliation{
	\institution{School of Computer Science and Center for OPTical IMagery Analysis and Learning (OPTIMAL), Northwestern Polytechnical University}
	\city{Xi'an} 
	\state{Shaanxi, P. R. China} 
	\postcode{710072}
}
\email{binzhao111@gmail.com}

\author{Xuelong Li}
\affiliation{
	\institution{Xi'an Institute of Optics and Precision Mechanics, Chinese Academy of Sciences}
	\city{Xi'an} 
	\state{Shaanxi, P. R. China} 
	\postcode{710019}
}
\email{xuelong_li@opt.ac.cn}

\author{Xiaoqiang Lu}
\affiliation{
	\institution{Xi'an Institute of Optics and Precision Mechanics, Chinese Academy of Sciences}
	\city{Xi'an} 
	\state{Shaanxi, P. R. China} 
	\postcode{710019}
}
\email{luxq666666@gmail.com}

\begin{abstract}
Exploiting the temporal dependency among video frames or subshots is very important for the task of video summarization. Practically, RNN is good at temporal dependency modeling, and has achieved overwhelming performance in many video-based tasks, such as video captioning and classification. However, RNN is not capable enough to handle the video summarization task, since traditional RNNs, including LSTM, can only deal with short videos, while the videos in the summarization task are usually in longer duration. To address this problem, we propose a hierarchical recurrent neural network for video summarization, called H-RNN in this paper. Specifically, it has two layers, where the first layer is utilized to encode short video subshots cut from the original video, and the final hidden state of each subshot is input to the second layer for calculating its confidence to be a key subshot. Compared to traditional RNNs, H-RNN is more suitable to video summarization, since it can exploit long temporal dependency among frames, meanwhile, the computation operations are significantly lessened. The results on two popular datasets, including the Combined dataset and VTW dataset, have demonstrated that the proposed H-RNN outperforms the state-of-the-arts.
\end{abstract}

\keywords{deep learning, video summarization, hierarchical recurrent neural network}

\maketitle

\section{Introduction}

Nowadays, video data are increasing explosively with the popularity of camera devices. There is a great demand for automatic techniques to handle these videos efficiently. Particularly, video summarization is one of the techniques that provide a viewer-friendly way to browse the huge amount of video data \cite{DBLP:conf/eccv/ZhangCSG16}. In general, it generates the video summary by shortening the video content into a compact version \cite{DBLP:conf/cvpr/YaoMR16}. Practically, there are several ways for video summarization to shorten the video. In this paper, we focus on the most popular one, i.e., key subshot selection.

\begin{figure}[t]
	\centering
	\includegraphics[width=0.45\textwidth]{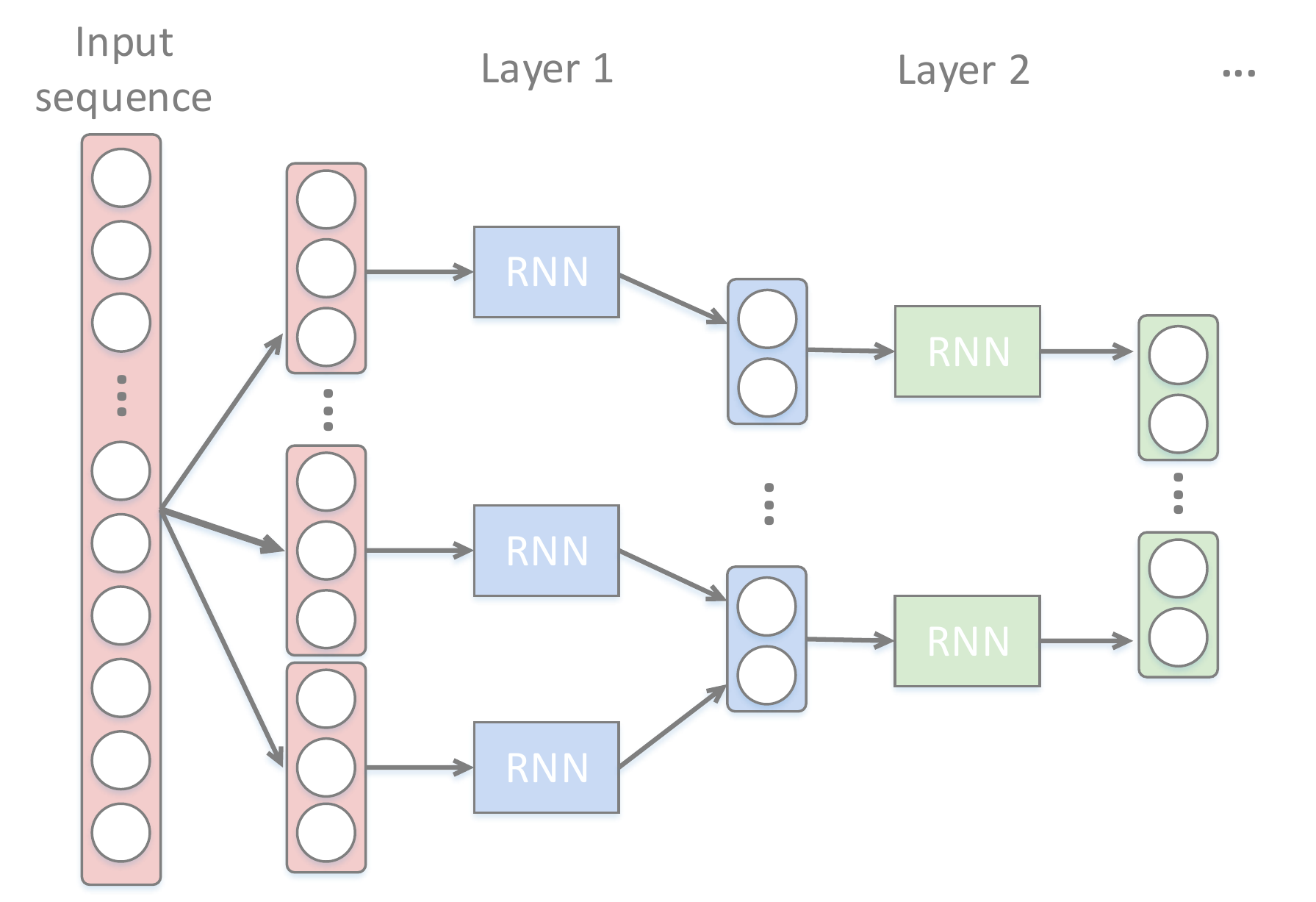}
	\caption{A compact architecture of hierarchical recurrent neural network. It is able to capture the long-range temporal dependency by processing the long input sequence hierarchically.}\label{Fig. 1}
\end{figure}

There is a stable growing interest in video summarization. Earlier works are mostly based on clustering or dictionary learning \cite{DBLP:conf/eccv/AnerK02,DBLP:journals/prl/AvilaLLA11,DBLP:conf/cvpr/ElhamifarSV12,DBLP:conf/cvpr/ZhaoX14a}, where the cluster centers or dictionary elements are taken as the most representative subshots in the video, and then selected into the summary as key subshots. Later, to emphasize the salient visual attributes in frames (people, objects, etc.), several property models are designed to capture the importance, representativeness and interestingness of the summary, and utilized to score the subshots \cite{DBLP:conf/eccv/PotapovDHS14,gygli2015video,lu2013story,DBLP:conf/eccv/GygliGRG14}. Naturally, those subshots with higher scores are selected into the summary. These approaches demonstrate promising results, and usually exceed the clustering and dictionary learning ones.

However, all these approaches are not capable enough to model the rich information in video content. Recently, inspired by the great success of deep learning, \emph{Convolutional Neural Network} (CNN) and \emph{Recurrent Neural Network} (RNN) are introduced to video summarization, where CNN is utilized to extract deep visual features and RNN is employed to predict the probability of one subshot to be selected into the summary \cite{DBLP:conf/cvpr/YaoMR16,DBLP:conf/eccv/ZhangCSG16}.  This architecture has achieved the state-of-the-art results in video summarization. Apart from the great ability of CNN in visual feature extraction, this mainly benefits from the capability of RNN in exploiting the temporal dependency among frames \cite{DBLP:conf/eccv/ZhangCSG16}. 

Unfortunately, RNN only works well for short frame sequence \cite{DBLP:conf/cvpr/NgHVVMT15,DBLP:conf/iccv/VenugopalanRDMD15}. Even for LSTM, one kind of RNN that is the most excellent in long frame sequence modeling, the favorable video length is less than 80 frames \cite{DBLP:conf/cvpr/NgHVVMT15}. While to video summarization, most of the videos contain thousands of frames. In this case, it is difficult for RNN to capture this long-range temporal dependency of videos. Thus, current approaches that apply RNN directly to video summarization may restrict the quality of video summary. To address this problem, we propose a hierarchical structure of RNN. As depicted in Figure 1, the hierarchical RNN is composed of multi-layers, and each layer is with one or more short RNNs, by which the long input sequence is processed hierarchically. Actually, the hierarchical RNN is a general architecture which varies according to specific tasks.

In this paper, a specialized hierarchical RNN is designed for the task of video summarization, called as H-RNN. Detailedly, it contains two layers. The first layer is a LSTM, which is utilized to process video subshots generated by cutting the video evenly, and the intra-subshot temporal dependency is encoded in the final hidden. Then, the final hidden of each subshot is input to the second layer. Specifically, the second layer is a bi-directional LSTM, which is composed of a forward and a backward LSTM. It is employed to exploit the inter-subshot temporal dependency and determine whether a certain subshot is valuable to be a key subshot. 

Generally, compared to current RNN-based approaches in video summarization, H-RNN has the following advantages:

1)	H-RNN can model long-range temporal dependency with a short time step. As a result, it reduces the information loss in frame sequence modeling meanwhile the computation operations are reduced significantly. 

2)	The hierarchical structure of H-RNN increases the nonlinear fitting ability of traditional RNN, which has been demonstrated extremely helpful for visual tasks \cite{DBLP:journals/corr/SimonyanZ14a,DBLP:conf/nips/SutskeverVL14}.

3)	H-RNN exploits the intra-subshot (i.e., among frames in the subshot) and inter-subshot temporal dependency in the two layers, respectively. This hierarchical structure is more suitable for video data, since video temporal structure is intrinsically layered as frames and subshots  \cite{DBLP:conf/cvpr/PanXYWZ16}.

\section{Related Works}

There have been a variety of video summarization approaches proposed in the literature. Generally, existing approaches can be classified into unsupervised ones and supervised ones.

Unsupervised approaches select key subshots according to manually designed criteria \cite{DBLP:conf/cvpr/ZhaoX14a,DBLP:conf/iccv/NgoMZ03,DBLP:conf/eccv/LiuK02,DBLP:journals/jodl/MundurRY06}, such as representative to the video content and diverse with each other, etc. Clustering is one of the most popular unsupervised summarization approaches \cite{DBLP:journals/jodl/MundurRY06,DBLP:conf/eccv/AnerK02,DBLP:journals/prl/AvilaLLA11}. Practically, with hand-crafted features, similar frames are grouped into the same cluster, and the cluster centers are taken as the most representative elements and selected into the summary. Earlier works apply clustering algorithms to video summarization directly \cite{DBLP:conf/icip/ZhuangRHM98,hadi2006video}. Later, more works integrate the domain knowledge of video data into clustering algorithms \cite{DBLP:journals/prl/AvilaLLA11,DBLP:journals/jodl/MundurRY06}. As in \cite{DBLP:journals/prl/AvilaLLA11}, the frames are initially clustered in sequential order, as consecutive frames are similar and more probably to be allocated to the same cluster. Other works construct more comprehensive models based on the idea of clustering \cite{DBLP:conf/iccv/NgoMZ03,DBLP:conf/cvpr/ChuSJ15}. As in \cite{DBLP:conf/iccv/NgoMZ03}, the video is transformed into an undirected graph, and the summary is generated by partitioning this graph into clusters. More recently, a co-clustering approach is proposed to simultaneously summarize several videos with the same topic by their co-occurrence, i.e., similar subshots shared by these videos are selected into the summary \cite{DBLP:conf/cvpr/ChuSJ15}.

Dictionary learning is another important unsupervised summarization approach \cite{DBLP:conf/cvpr/ElhamifarSV12,DBLP:conf/icmcs/LuanSLBLS14,DBLP:conf/cvpr/ZhaoX14a,DBLP:journals/tmm/CongYL12}. This kind of approaches seek to select a few key subshots to compose a compact dictionary so as to represent the video content. \cite{DBLP:conf/cvpr/ElhamifarSV12} supposes that the original video can be reconstructed by its summary sparsely. Based on this point, the summary is generated by sparse coding. Furthermore, the Locality-constrained Linear Coding (LLC) is introduced to \cite{DBLP:journals/tmm/LuWMGF14} to preserve the local similarity of video subshots when reconstructing the original video. Besides, to improve the efficiency, \cite{DBLP:conf/cvpr/ZhaoX14a} propose a quasi real-time dictionary learning approach to summarize the video, which updates the video summary on-the-fly by adding those elements that cannot be well reconstructed by current video summary. 

Supervised approaches learn the hidden summarization patterns from human generated summaries, which have been drawing increasing attention and getting more promising results than unsupervised ones \cite{DBLP:conf/cvpr/ZhangCSG16,DBLP:conf/nips/GongCGS14}. In supervised approaches, property models are usually taken as the decision criteria to select key subshots \cite{DBLP:conf/eccv/PotapovDHS14,DBLP:conf/cvpr/LeeGG12,lu2013story,DBLP:conf/eccv/GygliGRG14}. For example, \cite{DBLP:conf/cvpr/LeeGG12} and \cite{DBLP:conf/eccv/GygliGRG14} build importance model and interesting model to score the subshots, and those subshots with higher scores are selected into the summary. \cite{lu2013story} designs a stroyness model to constrain that the selected subshots have a smooth story line. Moreover, \cite{gygli2015video} employs three property models, i.e., interestingness, representativeness, uniformity, to build a comprehensive score function. Some other works even utilize auxiliary information to summarize videos, such as web image priors \cite{DBLP:conf/cvpr/KhoslaHLS13}, video titles \cite{song2015tvsum}, and video category labels \cite{DBLP:conf/eccv/PotapovDHS14}, etc.

\begin{figure*}[htp]
	\centering
	\subfigure[Hierarchical one-dimensional convolution]{
		\label{Fig. 2(a)}
		\includegraphics[width=0.34\textwidth]{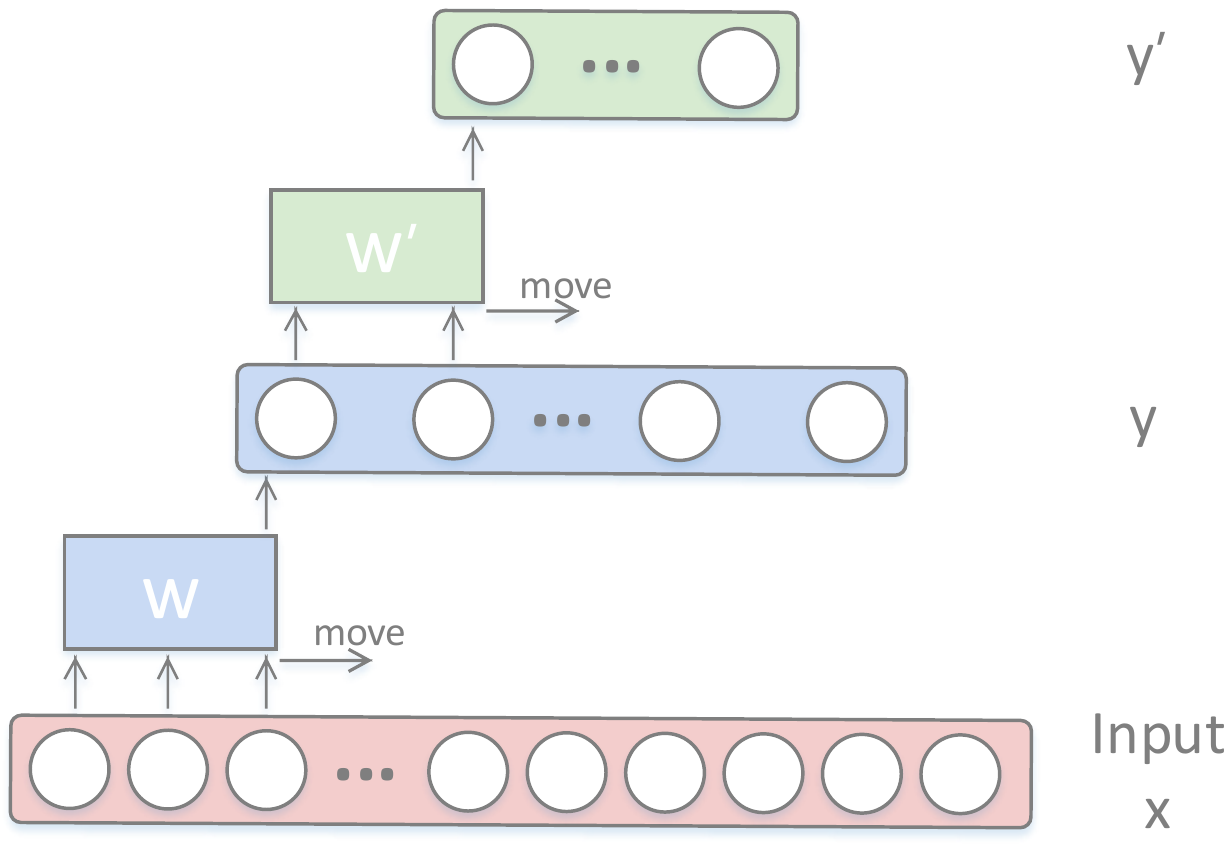}}
	\hspace{.7in}
	\subfigure[A single long recurrent neural network]{
		\label{Fig. 2(b)}
		\includegraphics[width=0.37\textwidth]{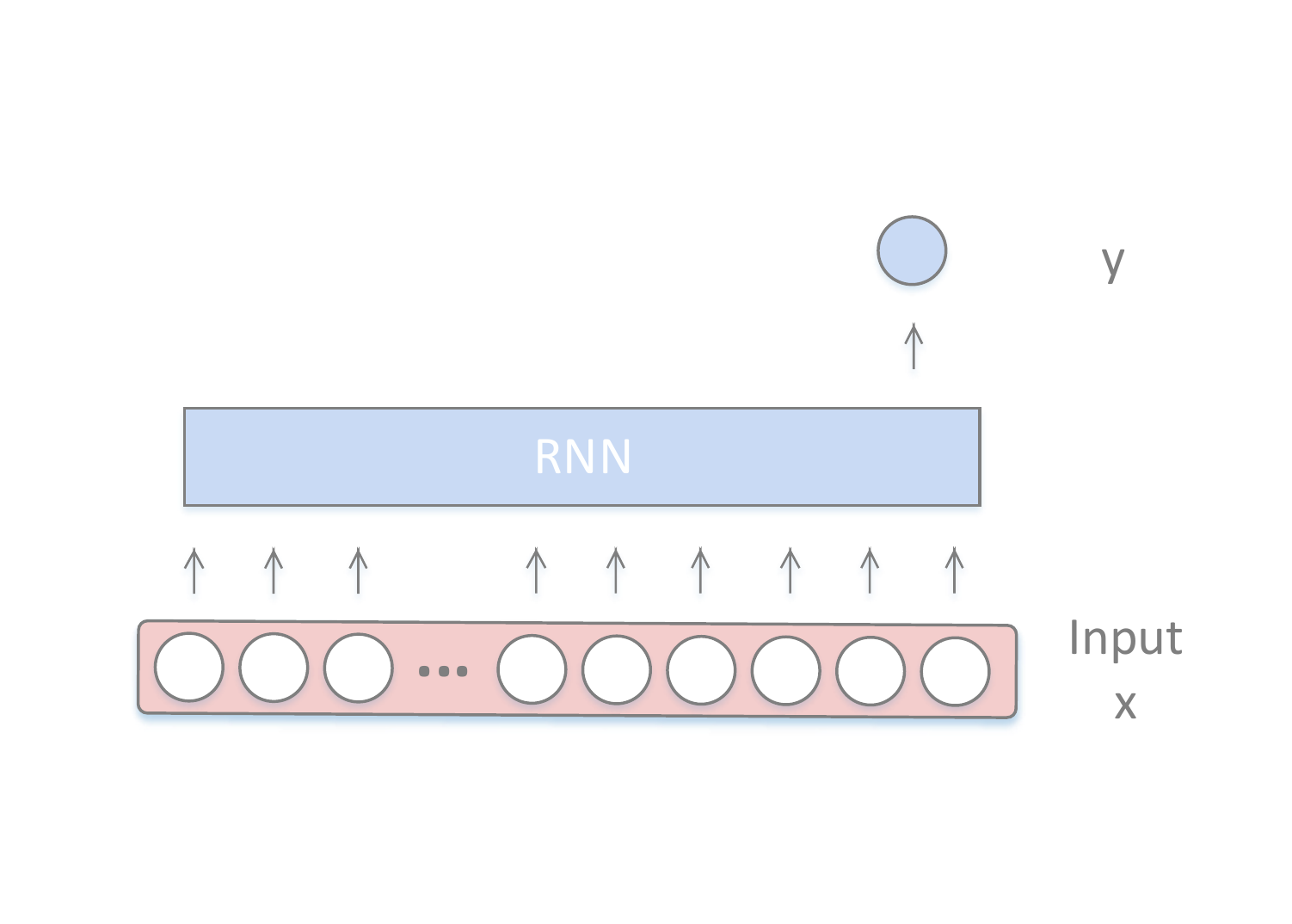}}
	\caption{The structures of hierarchical convolution and RNN. Actually, it becomes a comparison between hierarchical RNN and single long RNN, if we replace the filters in (a) (i.e., \(w\) and \(w'\)) with short RNNs. Compared to (b), hierarchical RNN is more efficient in long temporal dependency exploitation, meanwhile, the computation operations is significantly reduced.}
	\label{Fig. 2}
\end{figure*}

More recently, deep learning is introduced to video summarization, including CNN and RNN. \cite{DBLP:conf/cvpr/YaoMR16} builds a deep rank model relying on two CNNs .ie., AlexNet \cite{DBLP:conf/nips/KrizhevskySH12} and C3D \cite{DBLP:journals/corr/TranBFTP14} stitched with two Multi-Layer Perceptron (MLP) behind their final pooling layers. Given frames or subshots as input, the deep rank model outputs a ranking score. Naturally, a higher score indicates higher probability of that frame or subshot to be selected into the summary. Besides, LSTM is employed in \cite{DBLP:conf/eccv/ZhangCSG16,DBLP:conf/iccv/YangWLWGG15} to model the video sequence and rank the video subshots, which has achieved the state-of-the-art results in video summarization. However, due to the weakness in long temporal dependency exploitation, the input sequence to the LSTM is generated by the mean pooling or uniform sampling of frame features, which causes inevitable information loss. Actually, the proposed approach in this paper is essentially developed to solve this problem.

\section{Our approach}

In this section, we first provide a brief review to RNN, especially LSTM, since it is the build block of the proposed approach. Then, we present the hierarchical structure of RNN and our video summarization specialized hierarchical RNN, i.e., H-RNN.

\subsection{Recurrent Neural Network}

A standard RNN is constructed by extending a feedforward network with an extra feedback connection, so that it can model sequence. Practically, it can interpret the input sequence \(\left( {{x_1},{x_2}, \ldots ,{x_n}} \right)\) into another sequence \(\left( {{y_1},{y_2}, \ldots ,{y_n}} \right)\) iteratively by the following equations:
\begin{equation}{h_t} = \phi \left( {{W_h}{x_t} + {U_h}{h_{t - 1}} + {b_h}} \right),\end{equation}
\begin{equation}{y_t} = \phi \left( {{U_y}{h_t} + {b_y}} \right),\end{equation}
where \(h_t\) is the hidden state, \(t\) denotes the \(t\)-th time step, \(\phi \) stands for the activation function, and \(W\), \(U\) and \(b\) are the training weights and biases. 

Principally, the standard RNN should work efficiently in sequence modeling. However, it is really hard to train for the gradient vanishing problem \cite{bengio1994learning}. Then, LSTM is designed to address this issue, which is the most popular variant of standard RNN \cite{hochreiter1997long}. Specifically, it is extended from standard RNN with an extra memory cell, which is utilized to selectively memorize the previous inputs. In fact, there are several variants of LSTM, and they are similar with each other. In this paper, the one proposed in \cite{DBLP:journals/corr/ZarembaS14} is employed, which is most widely used in video-based tasks. Detailedly, the calculation of hidden state \(h_t\) and memory cell \(c_t\) is formulated as:

\begin{equation}{i_t} = \sigma \left( {{W_{ix}}{x_t} + {U_{ih}}{h_{t - 1}} + {b_i}} \right),\end{equation}
\begin{equation}{f_t} = \sigma \left( {{W_{fx}}{x_t} + {U_{fh}}{h_{t - 1}} + {b_f}} \right),\end{equation}
\begin{equation}{o_t} = \sigma \left( {{W_{ox}}{x_t} + {U_{oh}}{h_{t - 1}} + {b_o}} \right),\end{equation}
\begin{equation}{g_t} = \phi \left( {{W_{gx}}{x_t} + {U_{gh}}{h_{t - 1}} + {b_g}} \right),\end{equation}
\begin{equation}{c_t} = {f_t} \odot {c_{t - 1}} + {i_t} \odot {g_t},\end{equation}
\begin{equation}{h_t} = {o_t} \odot \phi \left( {{c_t}} \right),\end{equation}
where \(\sigma\) denotes the sigmoid function, and all the \(W\)s, \(U\)s, \(b\)s are the training weights and bias. Besides, \(i_t\), \(f_t\) and \(o_t\) are three gates, which are most important to LSTM. Concretely, the input gate \(i_t\) controls whether to record current input \(x_t\), the forget gate \(f_t\) decides whether to drop previous memory cell \(c_{t-1}\), and the output gate \(o_t\) determines the information in current memory cell \(c_t\) transfered to the hidden state \(h_t\).

\subsection{Hierarchical Recurrent Neural Network}

\begin{figure*}[t]
	\centering
	\includegraphics[width=0.95\textwidth]{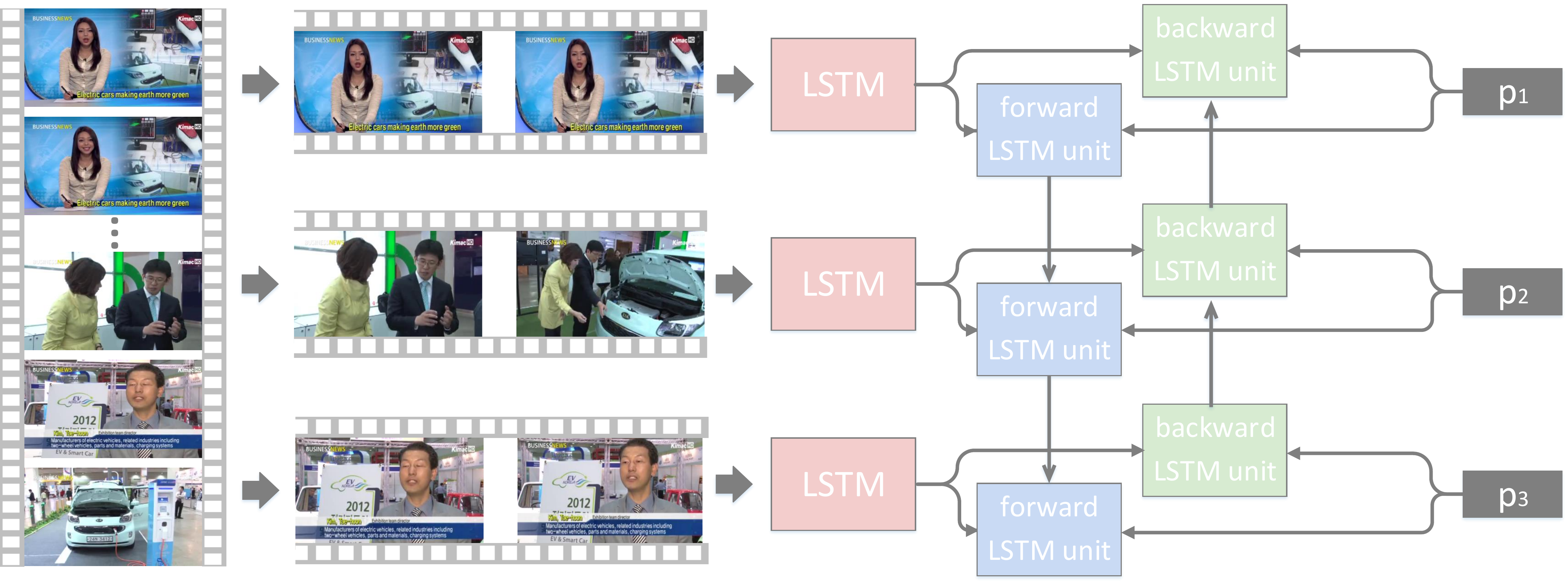}
	\caption{The architecture of the proposed approach for video summarization, i.e., H-RNN. It contains two layers, where the first layer is a LSTM and the second layer is a bi-directional LSTM (forward and backward). The two layers exploit the intra-subshot and inter-subshot temporal dependency, respectively, and the output of the second layer is utilized to predict the confidence of each subshot to be selected into the summary. }\label{Fig. 3}
\end{figure*}

The motivation for designing hierarchical RNN is to improve its capability to exploit long-range temporal dependency of the videos. Actually, it is originally inspired by the operation of one-dimensional convolution. As depicted in the first layer of Figure \ref{Fig. 2(a)}, a one-dimensional filter \(w\) is utilized to exploit the sequential information by performing convolutional operations on the input sequence \(x\):
\begin{equation}y = w * x,\end{equation}
where \(y\) denotes the output sequence, and \(*\) stands for the convolutional operation. It can be observed that although the filter \(w\) is much short than \(x\), at each time step, it operates appropriately on a subsequence of \(x\) and outputs a much shorter sequence \(y\). Particularly, if the convolution stride is set as \(n\), \(\left| y \right|\) is just \(1/n\) of \(\left| x \right|\), where \(\left|  \cdot  \right|\) denotes the length of the sequence. Furthermore, in the second layer, another filter \(w'\) is applied to \(y\), and outputs a shorter sequence \(y'\). Naturally, more filters can also be applied to higher layers, until the final output is generated. Then, the hierarchical structure is formed, and the long sequence \(x\) is processed by several short filters hierarchically.

Inspired by this, we construct a similar hierarchical structure for RNN. Actually, the filters in different layers of Figure \ref{Fig. 2(a)} are replaced by short RNNs, and the convolutional operation is just like successively processing several short subsequences which are cut from the long input sequence with or without overlap (the length of the subsequence is equal to the length of the filter RNN). Specifically, the RNN in the first layer exploits short-range temporal dependency, and longer ones are captured by higher layer RNNs. Intuitively, the long-range temporal dependency is captured by the hierarchical structure of several short RNNs. Moreover, compared to single RNN that operates on the long sequence directly in Figure \ref{Fig. 2(b)}, our hierarchical RNN can not only reduce the information loss in long sequence modeling, but also the computation operations are significantly lessened, which is detailedly discussed for the task of video summarization in the next subsection.

Generally, the hierarchical RNN can be composed of multi-layers and each layer with several RNNs. In other words, it is a general architecture that varies according to specific tasks.

\subsection{Video Summarization with H-RNN}

The videos for summarization are usually with long durations, i.e., about thousands of frames. Moreover, according to \cite{DBLP:conf/cvpr/PanXYWZ16}, the video structure is obviously layered that frames form the subshots and subshots form the video. Therefore, the hierarchical RNN is quite applicable to the task of video summarization. 

In this part, H-RNN is developed for the task of video summarization. As described in Figure \ref{Fig. 3}, it contains two layers, where the first layer is a LSTM and the second layer is a bi-directional LSTM. The details about summarizing the video with H-RNN is presented as follows.
  
Firstly, the frame sequence \(\left( {{f_1},{f_2}, \ldots ,{f_T}} \right)\) is separated into several subsequences, denoted as subshots \(\left( {{f_1},{f_2}, \ldots ,{f_s}} \right)\), \(\left( {{f_{s + 1}},{f_{s + 2}}, \ldots ,{f_{2s}}} \right)\) ,..., \(\left( {{f_{m \cdot s + 1}},{f_{m \cdot s + 2}}, \ldots ,{f_T}} \right)\), where \(f_i\) stands for the feature of frame \(i\), \(T\) denotes the total frames in the video, \(m\) is the number of subshots, and \(s\) is the length of each subshots. Practically, if the final subshot is shorter than \(s\), it is padded with zeros.

Then, the subshots are input to the first layer LSTM, which is formulated as follows:
\begin{equation}{\tau _i} = LSTM\left( {{f_{i \cdot s + 1}},{f_{i \cdot s + 2}}, \ldots ,{f_{\left( {i + 1} \right) \cdot s}}} \right),\end{equation}
where \(LSTM\left(  \cdot  \right)\) is short for Equ. (3)-(8), \({\tau _i}\) denotes the final hidden state of the \(i\)-th subshot. Actually, the short temporal dependency in the subshot is captured by \({\tau _i}\). Thus, it is taken as the representation of the \(i\)-th subshot.

Next, the sequence \(\left( {{\tau _1},{\tau _2}, \ldots ,{\tau _m}} \right)\) is input to the second layer. As aforementioned, the second layer is a bi-directional LSTM. Actually, bi-directional LSTM is composed of a forward LSTM and a backward LSTM. The main difference between them is that the backward LSTM operates reversely. Therefore, the calculation in the second layer is formulated as:
\begin{equation}h_t^f = LSTM\left( {{\tau _t},h_{t - 1}^f} \right),\end{equation}
\begin{equation}h_t^b = LSTM\left( {{\tau _t},h_{t + 1}^b} \right),\end{equation}
where \(h_t^f\) and \(h_t^b\) are the \(t\)-th output hidden state of forward LSTM and backward LSTM, respectively.

Finally, the output of the second layer is employed to predict the confidence of a certain subshot to be selected into the video summary. It is formulated as:
\begin{equation}{p_t} = softmax\left( {\tanh \left( {{W_p}\left[ {h_t^f,h_t^b,{\tau_t}} \right] + {b_p}} \right)} \right),\end{equation}
where \(W_p\) and \(b_p\) are the parameters to be learned. The softmax function is utilized to constrain the sum of the elements in \(p_t\) to be 1. Actually, \(p_t\) is a two-dimensional vector, each element of which indicates the possibility of the \(t\)-th subshot is key or non-key. It can be observed from Equ. (13) that \(p_t\) is determined jointly by the hidden state of forward and backward LSTM, i.e., \(h_t^f\) and \(h_t^b\), together with the representation of the \(t\)-th subshot \(\tau_t\). It is because that for subshot \(t\), \(h_t^f\) and \(h_t^b\) capture the front and behind temporal dependency, respectively, and \(\tau_t\) contains the intra-subshot dependency. All these information are quite important for the determination of whether to select subshot \(t\) into the summary.

In this paper, the proposed H-RNN is trained end-to-end. Given the reference summaries generated manually, the parameters in H-RNN are learned by:
\begin{equation}\Theta  = \mathop {\arg \min }\limits_\Theta  \frac{1}{N}\sum\limits_{i = 1}^N {\sum\limits_{t = 1}^{m^{(i)}} {L\left( {p_t^{\left( i \right)},g_t^{\left( i \right)}} \right)} }, \end{equation}
where \(N\) is the number of videos in the training set. \({m^{\left( i \right)}}\) denotes the number of shots in video \(i\). \(L\left(  \cdot  \right)\) stands for the loss function, which measures the cross-entropy between the generated probability distribution \({p_t^{\left( i \right)}}\) and the ground truth \({g_t^{\left( i \right)}}\). Practically, \({g_t^{\left( i \right)}}\) is a binary vector (indicates whether the subshot is key or not) or decimal vector (indicates the confidence of the subshot to be a key subshot).

Equ. (14) is optimized with \emph{Stochastic Gradient Descent} method based on \emph{Back Propagation Through Time} algorithm. Actually, H-RNN is easier to train than traditional LSTM, because the computation operations are much lessened \cite{DBLP:journals/pami/LiuHC10}. For example, for a video with 2000 frames, traditional LSTM needs 2000 operations to handle the whole frame sequence. While to H-RNN, the video is handled if the length of the first layer LSTM is 40 and the second layer bi-directional LSTM is 50, i.e., just 140 computation operations are needed totally. That is to say, more than ninety percent computation operations are reduced by H-RNN. 

\section{Experiments}

To verify the effectiveness of the proposed approach, it is tested on two popular dataset, i.e., Combined and VTW, and compared with several state-of-the-art approaches on video summarization.

\subsection{Setup}
\subsubsection{Dataset.}
The first dataset is combined with three popular datasets, i.e., SumMe \cite{DBLP:conf/eccv/GygliGRG14}, TVsum \cite{song2015tvsum} and MED \cite{DBLP:conf/eccv/PotapovDHS14}. In this paper, it is called the Combined dataset. The intuition lying behind the combination is that the videos in the three datasets are similar both in their visual contents and styles. Moreover, the combination of these datasets can address the problem of lacking of training data, which is widely used in video summarization. Detailedly, the Combined dataset consists of 235 videos, the average duration is 2 minutes, about 3000 frames for each video. In this paper, the Combined dataset is split into a training set of 180 videos, and a testing set of 55 videos.

The second dataset, i.e., VTW, is originally proposed for the task of video captioning, which totally contains 18100 videos \cite{DBLP:conf/eccv/ZengCNS16}. Fortunately, 2000 of them are labeled with subshot-level highlight scores that indicate the confidence of each subshot to be selected into the summary, so they are employed in this paper. Specifically, these videos are open-domain that crawled from YouTube, and the average duration is 1.5 minutes, about 2000 frames for each video. In this paper, the selected 2000 videos in the VTW dataset is divided into two parts, 1500 for training and 500 for testing.
\subsubsection{Feature.}

Both the shallow features and deep features are considered in this paper. Similar to prior works, for shallow features, color histogram, optical flow and SIFT features are extracted for each frame, they exploit the appearance, motion and local information, respectively \cite{song2015tvsum}. While for deep features, GoogLeNet is employed to extract the frame features, which is widely used in computer vision tasks \cite{DBLP:conf/cvpr/SzegedyLJSRAEVR15}.

\begin{table*}[t]
	\centering
	\caption{The results (F-measure) of various approaches on the Combined dataset. (The scores in bold indicate the best values.)}\label{Table1}
	\renewcommand\arraystretch{1.2}
	
	\begin{tabular}{c||p{1.5cm}<{\centering}|p{1.5cm}<{\centering}|p{1.5cm}<{\centering}||p{1.5cm}<{\centering}|p{1.5cm}<{\centering}|p{1.5cm}<{\centering}}
		\hline
		Feature &\multicolumn{3}{|c||}{shallow feature}&\multicolumn{3}{c}{deep feature}\\
		\hline	
		Datasets  &SumMe  & Tvsum  &MED&SumMe  & Tvsum  &MED  \\
		\hline
		VSUMM \cite{DBLP:journals/prl/AvilaLLA11} &0.328 &0.390&0.260&0.335 &0.391 &0.263\\
		LiveLight \cite{DBLP:conf/cvpr/ZhaoX14a} &0.357 &0.460&0.258&0.384 &0.477 &0.262  \\
		CSUV  \cite{DBLP:conf/eccv/GygliGRG14} &0.393 &0.532&0.277&-- & -- &--  \\
		LSMO \cite{gygli2015video} &0.397 &0.548&0.283&0.403 & 0.568 &0.285  \\
		Summary Transfer \cite{DBLP:conf/cvpr/ZhangCSG16} &0.397 &0.543&0.292&0.409 & 0.541 &0.297  \\
		vsLSTM  \cite{DBLP:conf/eccv/ZhangCSG16}   &0.406&0.571&0.288&0.421&0.580&0.293 \\
		dppLSTM \cite{DBLP:conf/eccv/ZhangCSG16} &0.407	&0.579	&0.294	&0.429	&0.597	&0.296	 \\
		\hline
		\hline
		H-RNN &\textbf{0.421}	&\textbf{0.602}	&\textbf{0.312}	&\textbf{0.443}	&\textbf{0.621}	&\textbf{0.311}\\
		\hline
		
	\end{tabular}
	
\end{table*}

\begin{figure*}[t]
	\centering
	\includegraphics[width=0.9\textwidth]{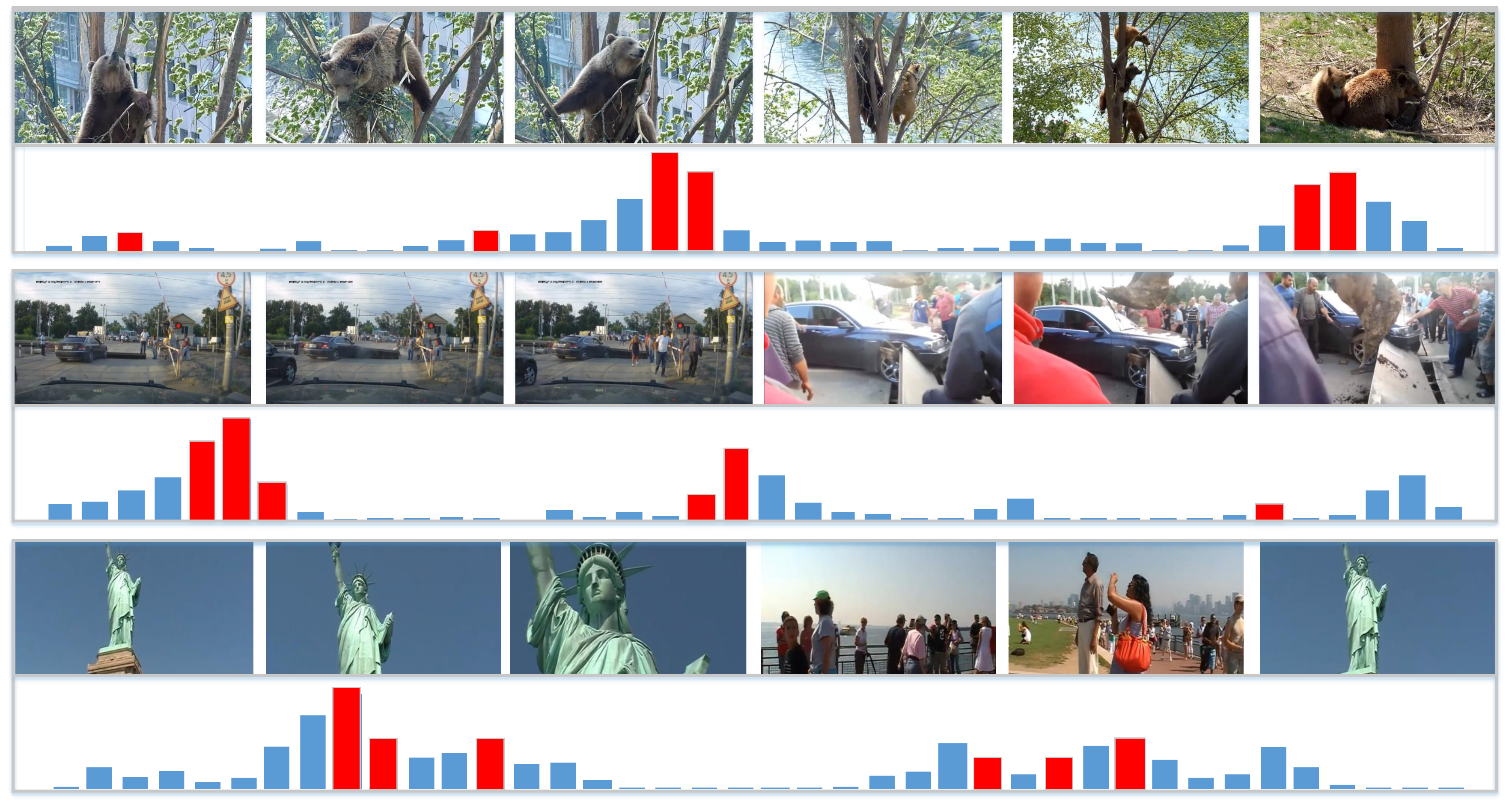}
	\caption{Three example summaries generated by H-RNN, where the key subshots in the summary are represented with several frames (one frame for each shot), and the histograms below the frames denote the distribution of human-annotated shot scores. The red histograms indicates the indexes of selected key subshots. Specifically, six key shots are displayed for each video, and they are corresponding to the upper frames sequentially.}\label{Fig. 4}
\end{figure*}

\subsubsection{Evaluation.}

As in prior works \cite{DBLP:conf/eccv/GygliGRG14,DBLP:conf/eccv/ZhangCSG16,gygli2015video}, the quality of the generated summary is evaluated by comparing to the reference summary created by human. There are three most frequently used evaluation metrics, i.e., precision (the correct subshots to subshots in the generated summary), recall (the correct subshots to subshots in the reference summary), F-measure (the harmonic mean of precision and recall), which are also employed in this paper.

\subsubsection{Parameters.}

We have tested several H-RNNs under different lengths of LSTMs in each layer, and they get stable performance when the LSTM length varies from 25-60 in each layer (40 is the most favorable). Therefore, in our H-RNN, the LSTM length of the first and second layer are both fixed as 40, so that the input flow at most goes through 80 steps. It can reduce the information loss caused by longer input flow.  As a result, our H-RNN can handle the frame sequence less than 1600. For videos contains more than 1600 frames, they are sampled to meet the constraint, while for videos with fewer than 1600 frames, they are padded with zeros.

\subsection{Results on the Combined dataset}

Table \ref{Table1} presents the performance of various approaches on the Combined dataset. It provides the results of different approaches on both shallow feature and deep feature. In this paper, for a fair comparison, these approaches are equipped with the same feature. Specifically, for all approaches, the shallow feature is generated by the combination of color histogram, optical flow and SIFT, and the deep feature is extracted by the pool5 layer of GoogLeNet. The two kinds of features are widely used in video summarization. From Table \ref{Table1}, it can be observed that most of the approaches perform better with deep feature rather than shallow feature, it benefits from the great capability of CNN to extract the visual information in the video.

\begin{table*}[t]
	\centering
	\caption{The results of various approaches on the VTW dataset.}\label{Table2}
	\renewcommand\arraystretch{1.2}
	
	\begin{tabular}{p{3cm}<{\centering}||p{1.5cm}<{\centering}|p{1.5cm}<{\centering}|p{1.5cm}<{\centering}||p{1.5cm}<{\centering}|p{1.5cm}<{\centering}|p{1.5cm}<{\centering}}
		\hline
		Feature &\multicolumn{3}{|c||}{shallow feature}&\multicolumn{3}{c}{deep feature}\\
		\hline	
		Metrics &Precision  & Recall  &F-measure&Precision & Recall &F-measure  \\
		\hline
		
		CSUV  \cite{DBLP:conf/eccv/GygliGRG14} &0.367 &0.423&0.399&-- & -- &--  \\
		HD-VS \cite{DBLP:conf/cvpr/YaoMR16} &-- &--&--&0.392 & 0.483 &0.427  \\
		
		vsLSTM  \cite{DBLP:conf/eccv/ZhangCSG16}   &0.388&0.490&0.433&0.397&0.495&0.446 \\
		\hline
		\hline
		H-RNN &\textbf{0.421}	&\textbf{0.522}	&\textbf{0.467}	&\textbf{0.432}	&\textbf{0.528}	&\textbf{0.478}\\
		\hline
		
	\end{tabular}
	
\end{table*}

\begin{table*}[t]
	\centering
	\caption{The results of various LSTM-based approaches on the VTW dataset.}\label{Table3}
	\renewcommand\arraystretch{1.2}
	
	\begin{tabular}{c||p{1.8cm}<{\centering}|p{1.8cm}<{\centering}|p{1.8cm}<{\centering}}
		\hline	
		Metrics &Precision  & Recall  &F-measure \\
		\hline
		single LSTM (mean pool)  &0.366 &0.442&0.387\\
		single LSTM (uniform sampling)  &0.336 &0.468&0.391\\
		bi-directional LSTM (mean pool) &0.383 &0.502&0.437  \\
		bi-directional LSTM (uniform sampling) &0.387 &0.495&0.431  \\
		\hline
		\hline		
		H-RNN (single)  &0.392 &0.523&0.455  \\
		H-RNN (bi-directional)  &\textbf{0.441} &\textbf{0.542}&\textbf{0.480}\\
		\hline

	\end{tabular}
	
\end{table*}

In Table \ref{Table1}, the compared approaches are from different types. The first five are non-RNN-based approaches. VSUMM and LiveLight summarize the video based on clustering and dictionary learning, respectively. CSUV and LSMO exceed them. Specifically, CSUV builds an \emph{Interstingness} model to predict the importance of each subshot. Moreover, LSMO is an extension of CSUV that combines several property models, including the \emph{Interstingness} model proposed in CSUV, together with \emph{Representativeness} model to constrain the key subshots to be representative to the video content, and \emph{Uniformity} model to demand key subshots distributing uniformly. Actually, the following two models are utilized to exploit the relationships between key subshots, i.e., temporal dependency. The better performance of LSMO than CSUV indicates that the temporal dependency is necessary for the task of video summarization. Summary Transfer also exploits the temporal dependency, meanwhile, it utilizes the category label of videos, which have achieved state-of-the-art results in non-RNN-based approaches. While the rest of the approaches are all RNN-based (i.e., LSTM). They performs better than non-RNN-based approaches even without auxiliary information, like the video category label in Summary Transfer. The better performance of RNN-based approaches has verified the superiority of LSTM in temporal dependency modeling.

Actually, vsLSTM and dppLSTM are two approaches most related to H-RNN. Concretely, vsLSTM is constructed by integrating a bi-directional LSTM with a \emph{Multi-Layer Perception} (MLP). Practically, it is reported that MLP is helpful in improving the summary quality \cite{DBLP:conf/cvpr/YaoMR16,DBLP:conf/eccv/ZhangCSG16}. However, it also increases the computation burden and the training parameters in the network. The proposed H-RNN outperforms vsLSTM even without the MLP, it benefits from the nonlinear fitting ability enhanced by the hierarchical structure of LSTM. Besides, dppLSTM is an extension of vsLSTM by adding a \emph{Determinatal Point Process} (DPP) model, which is proved effective in representative and diversity subset selection \cite{DBLP:conf/nips/GongCGS14,DBLP:conf/eccv/ZhangCSG16}. But the dppLSTM is much more complex than vsLSTM, also is hard to train. Generally, our H-RNN performs better than dppLSTM with a more compact architecture and with less computation. In conclusion, the better performance of H-RNN than vsLSTM and dppLSTM have verified the effectiveness and efficiency of H-RNN in the task of video summarization.

In Figure \ref{Fig. 4}, we present exemplar summaries generated by H-RNN. From the displayed key subshots and the human-annotated scores below them, it can be observed that most high score subshots are selected into our summaries, and the generated summaries can represent the original video content well. It indicates that, in most occasions, the summaries generated by H-RNN basically meet the human demand.

\subsection{Results on the VTW dataset}

Table \ref{Table2} shows the results of various approaches on the VTW dataset. Actually, many existing summarization approaches are based on manually designed criteria, so that they are quite dataset dependent. As a result, some of them are not suitable for the VTW dataset, and get very poor performance. For simplicity, they are not listed here. 

In Table \ref{Table2}, the results of three compared approaches are provided, where CSUV and vsLSTM have been introduced before. HD-VS summarizes the video by integrating two CNNs, i.e., AlexNet \cite{DBLP:conf/nips/KrizhevskySH12} and C3D \cite{DBLP:journals/corr/TranBFTP14}, together with two MLPs after the two CNNs. Specifically, AlexNet is employed to extract the visual information in each frame, and C3D is a 3D convolutional neural network that utilized to exploit the short-range temporal dependency. It can be observed from Table \ref{Table2} that RNN-based approaches, i.e., vsLSTM and H-RNN, show better results than HD-VS. It is because that LSTM does better in exploiting the temporal dependency than C3D, let alone in long frame sequence. Besides, the even better performance of H-RNN than vsLSTM also shows the superiority of the hierarchical structure of LSTM in video summarization.

To verify the necessity of the structure of H-RNN, the results of several approaches based on LSTM are listed in Table \ref{Table3}. Particularly, considering that LSTM does not work well with long frame sequence, the length of the compared approaches, i.e., single LSTM and bi-directional LSTM, are both fixed as 80. Limited by this, the frame features input to single LSTM and bi-directional LSTM are generated by the mean pooling or uniform sampling of the full frame feature sequence. It can be observed that the two frame feature treatment methods get comparable results. But the significantly better performance of bi-directional LSTM than single LSTM indicates that both the forward and backward temporal dependency are important for video summarization. It is also the reason that H-RNN (bi-directional) performs better than H-RNN (single), where H-RNN (bi-directional) denotes the second layer of H-RNN is a bi-directional LSTM, and the second layer of H-RNN (single) is a single LSTM. Besides, the better performance of H-RNN than single LSTM and bi-directional LSTM indicates that: 1) The hierarchical structure of H-RNN is more suitable for video summarization since it can deal with long frame sequence, while to single LSTM and bi-directional LSTM, they can only get satisfied results by mean pooling or uniformly sampling the frame feature sequence, which causes inevitable information loss. 2) The hierarchical structure of H-RNN increases the capability of non-linear fitting, which is helpful to the task of video summarization.

\section{Conclusions}

In this paper, we propose a hierarchical structure of RNN to enhance the capability of traditional RNN in long-range temporal dependency capturing. Particularly, for the task of video summarization, we design a specialized two-layer RNN according to the layered video structure, called as H-RNN. Particularly, the first layer is a LSTM, which is utilized to exploit the intra-subshot temporal dependency among frames. The second layer is a bi-directional LSTM that can capture both the forward and backward inter-subshot temporal dependency, and the output of the second layer is utilized to predict whether a certain subshot is valuable to be selected into the summary. Compared to current RNN-based approaches, H-RNN is more suitable to the task of video summarization, and the experimental results have verified its superiority.

\bibliographystyle{ACM-Reference-Format}
\bibliography{sigproc} 

\end{document}